\documentclass{article}

\PassOptionsToPackage{numbers}{natbib}

 \usepackage[dblblindworkshop, final]{neurips_2025}
  \usepackage{graphicx}
  \usepackage{subcaption}
  \usepackage{booktabs}
\workshoptitle{First Workshop on CogInterp: Interpreting Cognition in Deep Learning Models}

\usepackage[utf8]{inputenc} % allow utf-8 input
\usepackage[T1]{fontenc}    % use 8-bit T1 fonts
\usepackage{hyperref}       % hyperlinks
\usepackage{url}            % simple URL typesetting
\usepackage{booktabs}       % professional-quality tables
\usepackage{amsfonts}       % blackboard math symbols
\usepackage{nicefrac}       % compact symbols for 1/2, etc.
\usepackage{microtype}      % microtypography
\usepackage{xcolor}         % colors

\title{Do Sparse Subnetworks Exhibit Cognitively Aligned Attention? Effects of Pruning on Saliency Map Fidelity, Sparsity, and Concept Coherence}

\author{%
  Sanish Suwal\thanks{Equal contribution.}\\
  Rochester Institute of Technology\\
  Rochester, NY \\
  \texttt{ss4657@rit.edu } \\
  \And
  Dipkamal Bhusal\footnotemark[1]\\
  Rochester Institute of Technology\\
  Rochester, NY \\
  \texttt{db1702@rit.edu } \\
  \AND
  Michael Clifford \\
Toyota InfoTech Labs\\
   Mountain View, CA \\
  \texttt{michael.clifford@toyota.com} \\
  \And
  Nidhi Rastogi \\
  Rochester Institute of Technology\\
  Rochester, NY \\
  \texttt{nxrvse@rit.edu} \\
}

\begin{document}

\maketitle

\begin{abstract}
Prior works have shown that neural networks can be heavily pruned while preserving performance, but the impact of pruning on model interpretability remains unclear. In this work, we investigate how magnitude-based pruning followed by fine-tuning affects both low-level saliency maps and high-level concept representations. Using a ResNet-18 trained on ImageNette, we compare post-hoc explanations from Vanilla Gradients (VG) and Integrated Gradients (IG) across pruning levels, evaluating sparsity and faithfulness. We further apply CRAFT-based concept extraction to track changes in semantic coherence of learned concepts. Our results show that light-to-moderate pruning improves saliency-map focus and faithfulness while retaining distinct, semantically meaningful concepts. In contrast, aggressive pruning merges heterogeneous features, reducing saliency map sparsity and concept coherence despite maintaining accuracy. These findings suggest that while pruning can shape internal representations toward more human-aligned attention patterns, excessive pruning undermines interpretability.  Code is available at \url{https://github.com/sanishsuwal7/Neurips-CogInterp/}.
\end{abstract}

\section{Introduction}
Humans often rely on a sparse subset of cues to make decisions, focusing attention on the most salient and semantically meaningful aspects of a scene. In contrast, deep neural networks are often criticized for relying on hard-to-interpret features. Neural network pruning is a widely studied approach for removing unnecessary weights or structure from overparameterized models~\cite{han2015learning, he2018soft}. By eliminating low-magnitude weights, pruning can drastically reduce model size and computation with minimal accuracy loss. In some cases, models like ResNet-50 can be pruned by 80–90\% and still retain performance comparable to their dense counterparts~\cite{frankle2018lottery}, suggesting that much of a trained network’s capacity may be redundant.

While pruning’s impact on efficiency is well understood, its effect on interpretability is far less explored. In high-stakes domains such as healthcare or autonomous driving, models must not only be accurate but also explainable in a way that aligns with human reasoning. Unfortunately, post-hoc saliency methods often produce noisy, unfaithful explanations that misrepresent the model's actual decision process~\cite{adebayo2018sanity, bhusal2023sok}.

In this work, we investigate how pruning affects interpretability across two levels of abstraction: \textbf{a) Low-level attribution maps}, evaluated in terms of sparsity: \textit{the concentration of attribution on a small set of relevant input features}~\cite{chalasani2020conciseexplanationsneuralnetworks}, and faithfulness: \textit{how accurately explanations reflect the features that influence predictions}~\cite{rong22consistent}. \textbf{b) High-level concept representations}, assessed qualitatively by examining how the most important extracted concepts change in appearance, and coherence across pruning levels, using concept extraction method such as CRAFT~\cite{fel2023craft}.

We hypothesize that pruning acts as a structural regularizer, eliminating redundant pathways and forcing the model to rely on a smaller, more essential set of discriminative features. This, in turn, could yield more focused saliency maps and cleaner, semantically distinct concepts. However, aggressive pruning may compress multiple discriminative features into fewer activation patterns, reducing concept coherence even if low-level sparsity improves.

To test these hypotheses, we conduct a systematic study using a ResNet-18~\cite{he2016deep} trained on ImageNette~\cite{imagenette}. We apply global magnitude pruning with iterative fine-tuning following the lottery ticket hypothesis framework~\cite{frankle2018lottery}, and evaluate post-hoc explanations using Vanilla Gradients (VG)\cite{simonyan2014deepinsideconvolutionalnetworks} and Integrated Gradients (IG)\cite{sundararajan2017axiomaticattributiondeepnetworks}. At the concept level, we use CRAFT~\cite{fel2023craft} to measure qualitative changes in semantic coherence of discovered concepts.

Our results show that light-to-moderate pruning improves saliency-map focus and faithfulness while preserving distinct, semantically meaningful concepts. In contrast, aggressive pruning blurs concept boundaries, merging heterogeneous visual patterns despite maintaining accuracy. These findings suggest that appropriate amount of pruning can shape internal representations toward more human-aligned attention, but excessive pruning undermines the quality of learned concepts, highlighting a nuanced trade-off between sparsity, faithfulness, and semantic coherence.

\section{Related Work}
The Lottery Ticket Hypothesis~\cite{frankle2018lottery} showed that sparse subnetworks can be retrained to match full-model accuracy. Later, \citet{frankle2019dissecting} used Network Dissection to find that heavy pruning of ResNet-50 preserves most human-recognizable concepts. \citet{hooker2019compressed} noted that compressed models may forget certain examples, though they did not assess explanation quality. \citet{weber2023less} observed reduced noise in GradCAM maps after moderate pruning in VGG-16, but without quantitative rigor or fine-tuning. \citet{tan2024evaluating} reported that pruning without fine-tuning can collapse explanations despite stable predictions. \citet{suwal2025smaller} studied the impact of adversarial training and pruning on saliency maps of vehicular datasets. Our work differs by quantitatively evaluating post-hoc explanations under magnitude pruning with fine-tuning, using ROAD and Gini metrics, and extending analysis to concept-level changes, using automatic concept extraction method, CRAFT~\cite{fel2023craft}.

\section{Methodology}
We study how iterative magnitude-based pruning and fine-tuning affect both pixel-level explanations (saliency maps) and high-level concept representations. Our base model is ResNet-18~\cite{he2016deep} trained on the ImageNette dataset~\cite{imagenette}, a 10-class subset of ImageNet \cite{deng2009imagenet} designed for fast benchmarking.

\begin{table}[h]
\centering
\caption{Model performance on ImageNette test-set across several pruning levels}
\label{tab:accuracy}
\resizebox{0.60\textwidth}{!}{%
\begin{tabular}{@{}lllllll@{}}
\toprule
\textbf{Pruning \%} & 0     & 10    & 20    & 30    & 50    & 70    \\ \midrule
\textbf{Accuracy}   & 84.15 & 83.08 & 84.31 & 85.58 & 84.99 & 83.99 \\ \bottomrule
\end{tabular}%
}
\end{table}

\subsection{Pruning Strategy}
The Lottery Ticket Hypothesis~\cite{frankle2018lottery} shows that, within a randomly initialized, dense neural network, there exist sparse subnetworks, that when trained in isolation from their original initialization, can match or exceed the accuracy of the full network in the same number of training iterations. A winning ticket is thus a subnetwork whose initial weights and structure are sufficient to learn the target task without relying on the excess parameters in the original model.

Following this framework, we start with a ResNet-18~\cite{he2016deep} model randomly initialized with parameters, $\theta$, and train it to convergence on the ImageNette dataset~\cite{imagenette}. After training, we perform global unstructured magnitude pruning to remove a fixed proportion $p$\% of the smallest-magnitude weights across all layers (excluding biases), producing a binary pruning mask $M$. The surviving weights $\theta \cdot M$ are then reset to their original initialization values from $\theta$, yielding a winning ticket candidate, which is fine-tuned for preserving model accuracy. We repeat the prune–fine-tune cycle for $n$ iterations, progressively increasing weight-sparsity while maintaining high predictive performance. 

Classification accuracy on the ImageNette test set remains within 1–2 percentage points of the unpruned baseline up to 70\% weight removal (Table \ref{tab:accuracy}), consistent with lottery ticket hypothesis findings that substantial weight-sparsity can be introduced without significant performance loss.

\subsection{Post-hoc explanation and concept extraction}

For pixel-level explanations, we generate saliency maps using Vanilla Gradients (VG)~\cite{simonyan2014deepinsideconvolutionalnetworks} and Integrated Gradients (IG)~\cite{sundararajan2017axiomaticattributiondeepnetworks}. We quantify their quality using two metrics:
\textbf{1) Sparsity}~\cite{chalasani2020conciseexplanationsneuralnetworks}, measured via the Gini coefficient, which captures the concentration of attribution values in a small set of pixels. Higher values indicate sparse and comprehensible saliency maps. \textbf{2) Faithfulness}, measured using ROAD MoRF strategy~\cite{rong22consistent}. Features are progressively removed in decreasing order of importance, and model accuracy is recorded at each step. Unlike Insertion/Deletion~\cite{petsiukrise} or ROAR~\cite{hooker2019benchmark}, ROAD avoids distribution shifts from synthetic perturbations and does not require retraining. We also measure faithfulness using the area over the perturbation curve (AOPC), where higher values correspond to more accurate identification of critical features. Further metric details are provided in Appendix~\ref{appendix:metrics}.

For high-level concept-level analysis, we adopt the CRAFT pipeline~\cite{fel2023craft} to extract the top-ranked concepts for selected classes at different pruning stages. We use a patch size of 64 and concept extraction of 10 ranks (default in the official implementation), and follow the publicly available code\footnote{\url{https://github.com/deel-ai/Craft}}. We then qualitatively compare the discovered concepts before and after pruning, noting changes in semantic composition. Further details on CRAFT is provided in Appendix~\ref{appendix:craft}.

\section{Results}
\subsection{Saliency maps}
\begin{figure}[htbp]
    \centering
    \begin{subfigure}[b]{0.45\textwidth}
        \includegraphics[width=\linewidth]{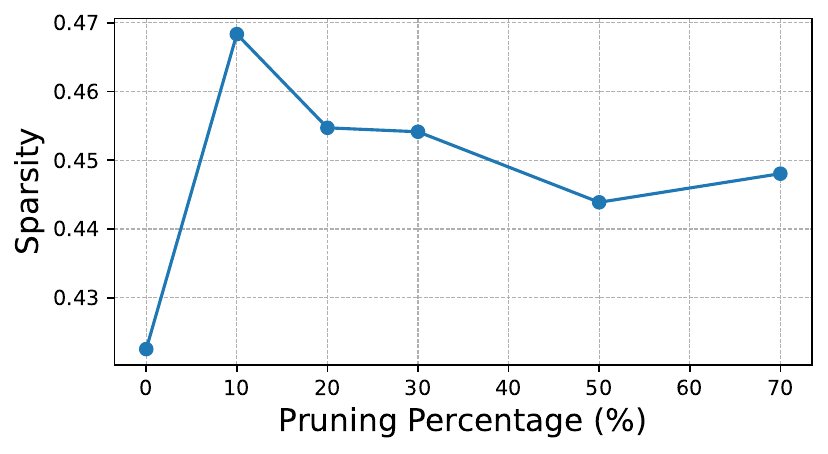}
        \caption{Vanilla Gradient (VG)}
        \label{fig:sparse1}
    \end{subfigure}
    \hfill % Adds horizontal space between subfigures
    \begin{subfigure}[b]{0.45\textwidth}
        \includegraphics[width=\linewidth]{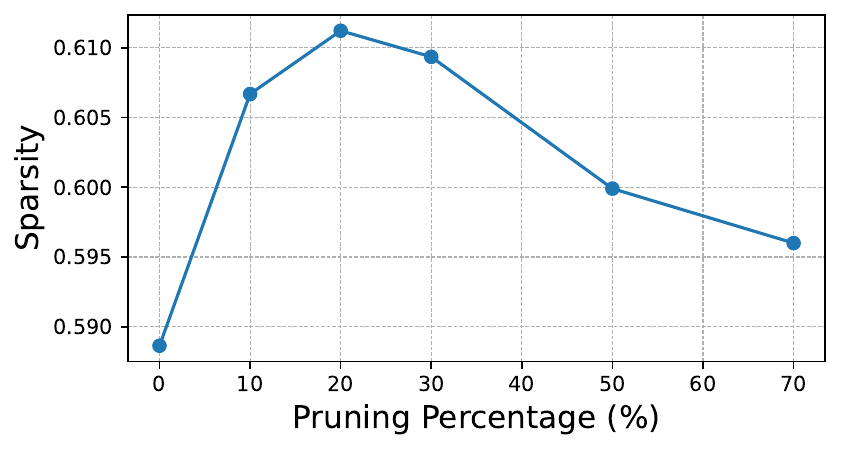}
        \caption{Integrated Gradients (IG)}
        \label{fig:sparse2}
    \end{subfigure}
    \caption{Sparsity evaluation of saliency maps.}
    \label{fig:sparsefigure}
\end{figure}

\textbf{Sparsity.} Figure~\ref{fig:sparsefigure} shows saliency sparsity (Gini index) for Vanilla Gradients (VG) and Integrated Gradients (IG) as pruning increases. For both methods, sparsity rises with pruning, indicating more concentrated attributions in pruned models. VG peaks near $\sim$10\% pruning and IG near $\sim$20\%, after which improvements plateau. Across all levels, IG maps are consistently \emph{absolutely} sparser than VG, reflecting IG’s built-in tendency to concentrate attribution scores.

\begin{figure}[htbp]
    \centering
    \begin{subfigure}[b]{0.35\textwidth}
        \includegraphics[width=\linewidth]{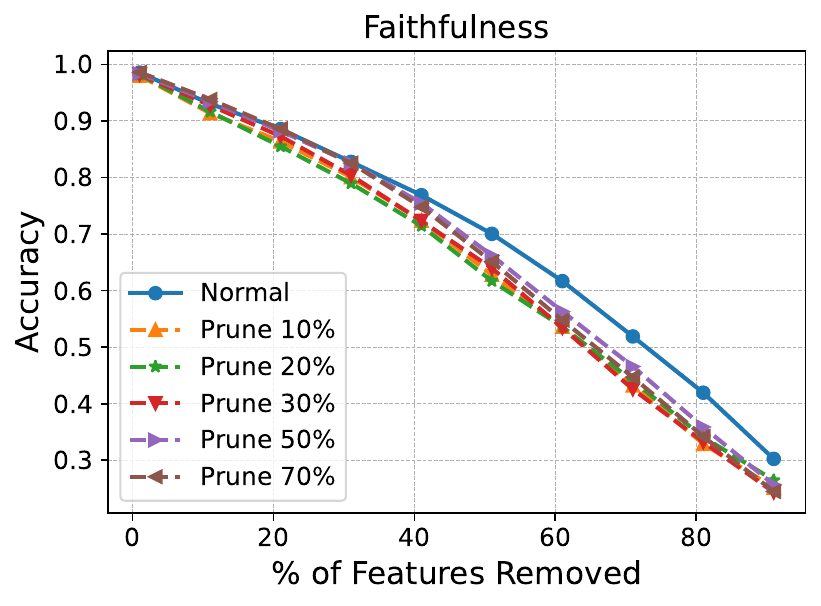}
        \caption{Vanilla Gradient (VG)}
        \label{fig:subim1}
    \end{subfigure}
    \hfill % Adds horizontal space between subfigures
    \begin{subfigure}[b]{0.35\textwidth}
        \includegraphics[width=\linewidth]{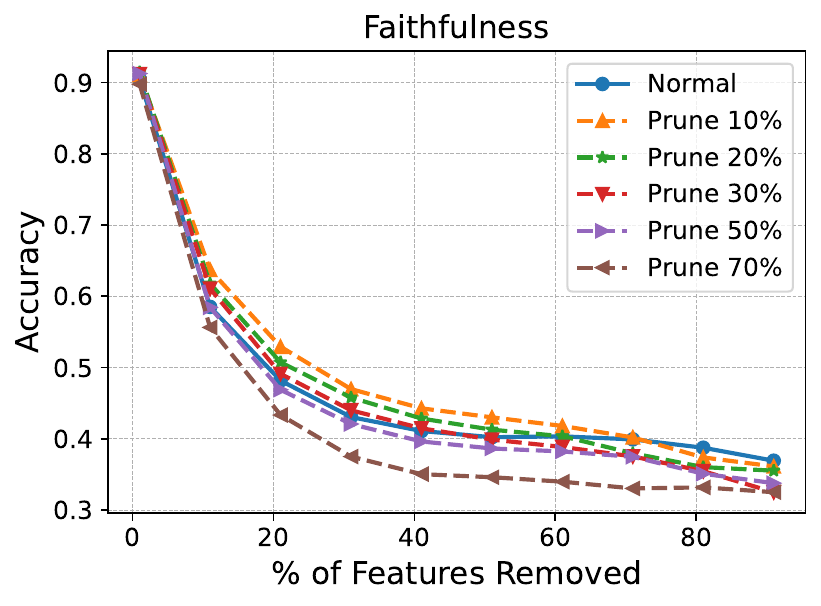}
        \caption{Integrated Gradients (IG)}
        \label{fig:subim2}
    \end{subfigure}
    \caption{Faithfulness evaluation. Sharper drop in model accuracy signifies faithful saliency maps.}
    \label{fig:roadplot}
\end{figure}

\begin{figure}[htbp]
    \centering
    \begin{subfigure}[b]{0.35\textwidth}
        \includegraphics[width=\linewidth]{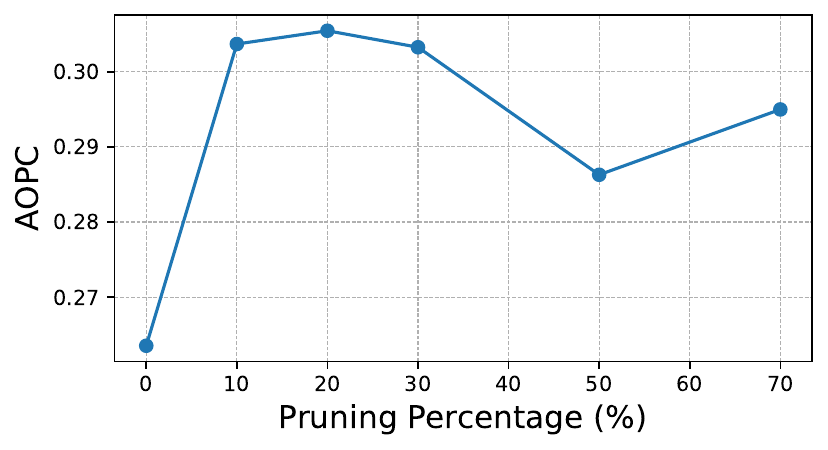}
        \caption{Vanilla Gradient (VG)}
        \label{fig:aopc1}
    \end{subfigure}
    \hfill % Adds horizontal space between subfigures
    \begin{subfigure}[b]{0.35\textwidth}
        \includegraphics[width=\linewidth]{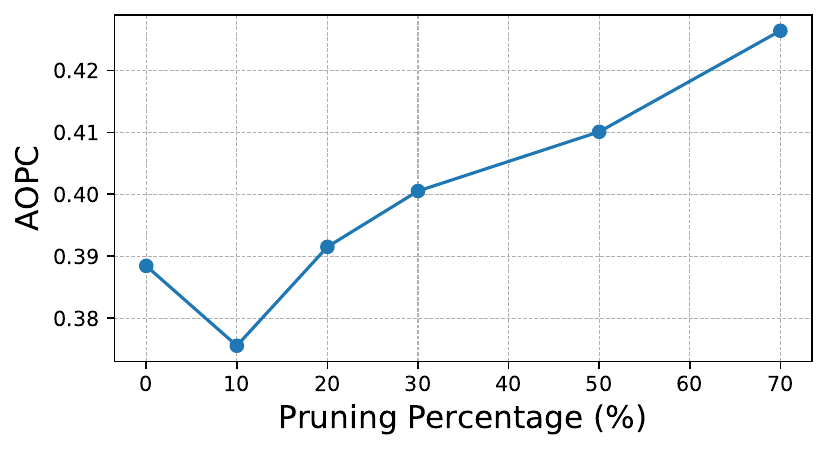}
        \caption{Integrated Gradients (IG)}
        \label{fig:aopc2}
    \end{subfigure}
    \caption{Faithfulness evaluation using AOPC of ROAD-MoRF plots.}
    \label{fig:aopc}
\end{figure}

\textbf{Faithfulness.} We assess faithfulness with ROAD–MoRF curves (Fig.~\ref{fig:roadplot}) and summarize with AOPC scores (Fig.~\ref{fig:aopc}). A steeper accuracy drop under MoRF indicates that removed, high-ranked features were truly critical.

\emph{Vanilla Gradients (VG).} Pruning produces \emph{sharper} ROAD curves than the natural model across most levels, with clear gains already at 10–20\% pruning. Additional pruning maintains or slightly reduces these gains. AOPC mirrors this trend: it peaks at 20\% and remains above baseline thereafter.

\emph{Integrated Gradients (IG).} Faithfulness is largely unchanged at $\leq$20\% pruning: the curves closely track the unpruned model, and early drops can be slightly shallower. Improvements emerge beyond 30\%, with the largest gains at heavy pruning (50–70\%), where curves are steepest and AOPC reaches its maximum.

\textbf{Takeaway.} Pruning’s faithfulness gains are \emph{method-dependent}: VG benefits at light-to-moderate sparsity (peaking at $\sim$20\%), whereas IG requires moderate to higher pruning. 

%For VG, pruning consistently improves explanation faithfulness relative to the unpruned model, as indicated by sharper accuracy drops when top-ranked features are removed. Gains are visible even at light pruning levels (10–20\%), with the Prune 20\% model showing the steepest early decline. Beyond moderate sparsity, additional pruning maintains or slightly reduces faithfulness, suggesting that VG benefits most from moderate weight removal. This is corroborated by the AOPC scores, which peak at 20\% pruning and gradually decline while remaining above baseline for all pruning levels.

%In contrast, IG exhibits minimal faithfulness change at light pruning levels ($\leq$ 20\%), with curves closely tracking the unpruned model and in some cases showing slightly shallower initial drops. Improvements become apparent only beyond 30\% pruning, with the largest gains at 70\% pruning, where the ROAD curves show the steepest declines and AOPC values reach their maximum. This pattern suggests that IG requires substantial sparsity before its attributions align more strongly with truly predictive features.

%Overall, the results indicate that pruning’s impact on explanation faithfulness is method-dependent: VG is sensitive to even small reductions in model size, while IG demands heavier pruning to achieve substantial gains.

%These results suggest that, IG benefits with only heavy pruning whereas Vanilla Gradient explanations may benefit from pruning across a broader range of pruning levels.

\textbf{Qualitative maps.} Visual inspection (provided in Appendix~\ref{appendix:qualitative}) aligns with these patterns: 10–30\% pruning sharpens object focus for both VG and IG; $\geq$50\% pruning begins to reintroduce background noise and reduces the apparent sparsity gains.

\subsection{Concept analysis under pruning}
We complement pixel-level metrics with a concept-level probe using CRAFT~\cite{fel2023craft} on the ``parachute'' class (Appendix~\ref{appendix:concept}). The unpruned model’s top concepts are highly distinctive and semantically pure, dominated by brightly colored canopy sections against clear skies, with secondary cues such as fabric color blocks. Light pruning (10–20\%) largely preserves these object-centric patterns, concentrating importance on zoomed-out parachute shots and bold canopy stripes while reducing the variety of environmental cues. From 30\% pruning onward, concept coherence declines: object-relevant features increasingly co-occur with unrelated textures, water scenes, or fabric folds, and background reliance becomes more pronounced.  By 70\% pruning, the top concept still contains parachute cues, but most remaining concepts are mixed or irrelevant, including abstract textures, text, and structural imagery, indicating a substantial loss of semantic clarity. 

These results suggest that moderate pruning can reweight the model toward high-confidence object cues while retaining coherent concept structure, but heavy pruning forces disparate visual patterns into fewer activation clusters, degrading interpretability despite occasional improvements in saliency sparsity. 

\section{Conclusion}
This work examined how structured magnitude pruning influences the quality of post-hoc saliency explanations and concept-based interpretations, focusing on sparsity, faithfulness, and coherence. Across ImageNette experiments with ResNet-18, we observed that moderate pruning increases saliency sparsity and faithfulness, and compact concept activations.  These results suggest pruning not only compresses models but can also improve post-hoc explanations.

\section*{Acknowledgment}
This work was supported by Toyota InfoTech Labs through Unrestricted Research Funds.

\bibliographystyle{plainnat}
\bibliography{reference}

\appendix
\newpage 
\section{Qualitative result}\label{appendix:qualitative}

\begin{figure}[h]
    \centering
    \includegraphics[width=0.85\linewidth]{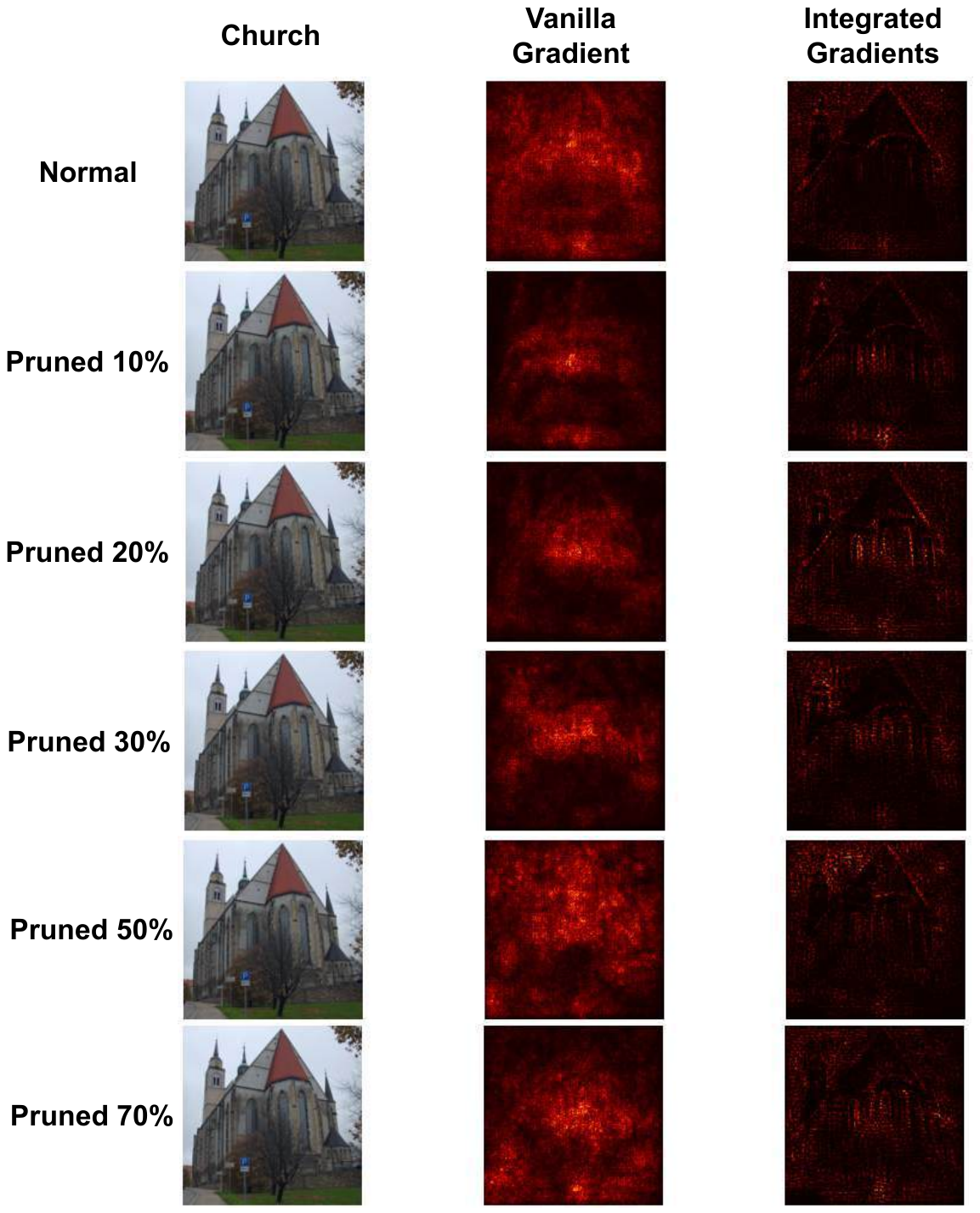}
    \caption{Church}
    \label{fig:church}
\end{figure}

\begin{figure}[h]
    \centering
    \includegraphics[width=0.85\linewidth]{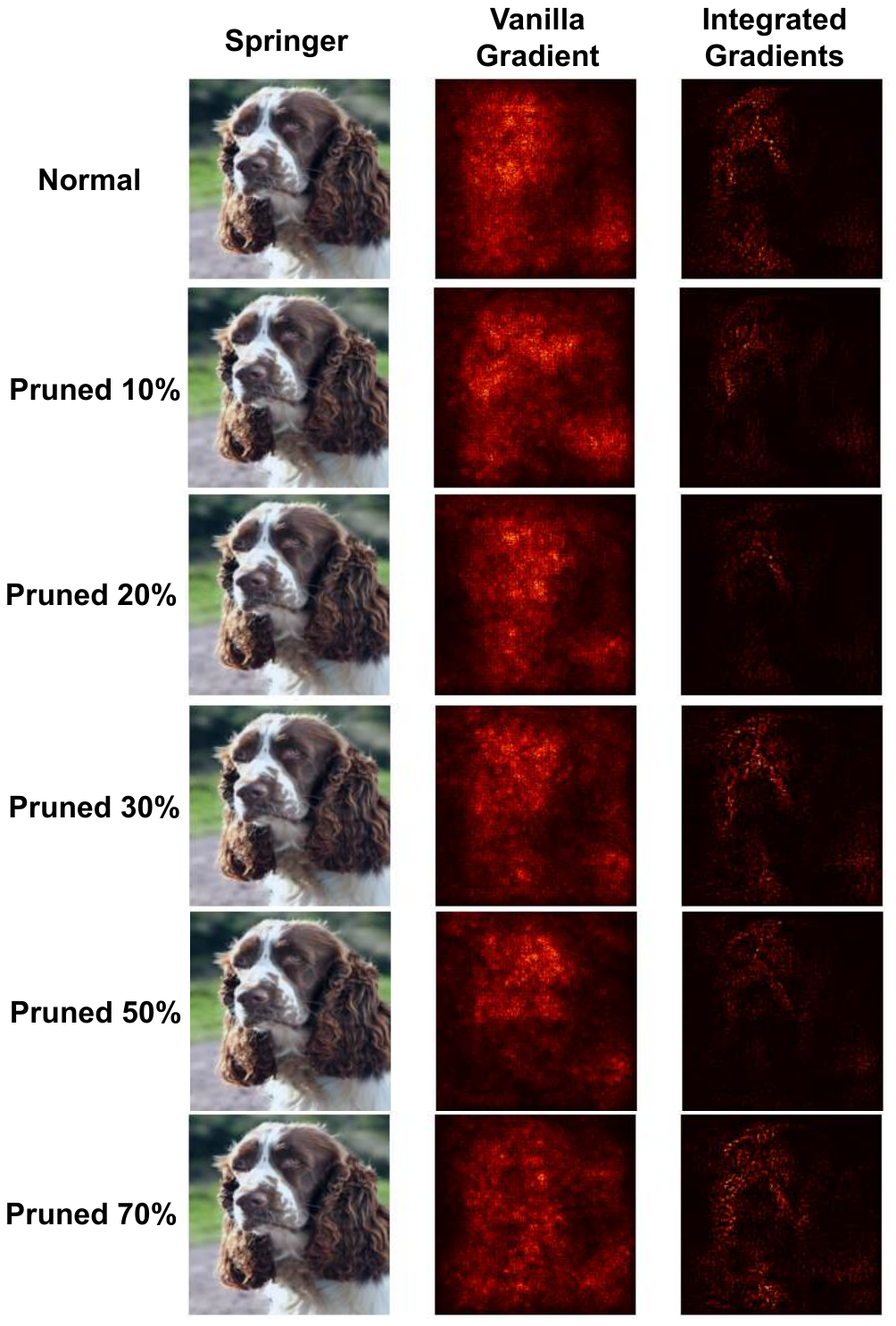}
    \caption{Springer}
    \label{fig:springer}
\end{figure}

\begin{figure}
    \centering
    \includegraphics[width=0.85\linewidth]{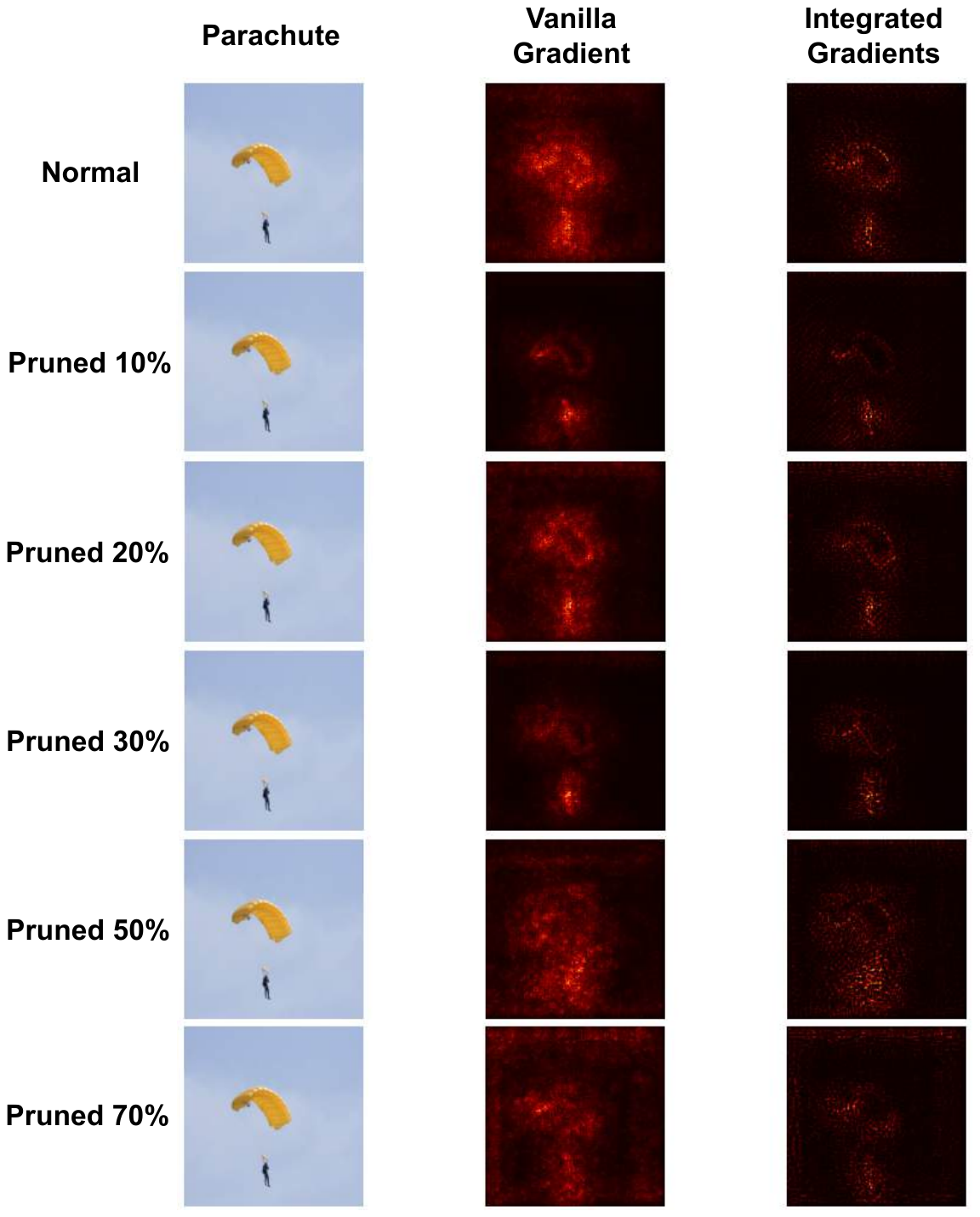}
    \caption{Parachute}
    \label{fig:parachute}
\end{figure}

Figures~\ref{fig:church}, \ref{fig:springer}, and \ref{fig:parachute} present Vanilla Gradient (VG) and Integrated Gradients (IG) saliency maps for the normal and pruned models. Across all examples, VG tends to produce noisier and less focused explanations compared to IG, but pruning influences both methods.

In Church (Fig. \ref{fig:church}), with VG, noise is prominent for the normal model and persists after pruning, though slight visual sharpening occurs up to 30\% pruning. IG produces clearer focus on the frontal architecture in the pruned models; however, at 50–70\% pruning, maps become noisier, aligning with the sparsity drop seen in quantitative results.

In Springer (Fig. \ref{fig:springer}), VG explanations remain diffuse and noisy across all pruning levels. IG maps, however, become sharper and more focused up to 30\% pruning, before degrading in clarity at higher pruning levels. This mirrors the trend where pruning beyond a moderate threshold reduces sparsity and comprehensibility.

In Parachute (Fig. \ref{fig:parachute}), VG maps start noisy and improve moderately with pruning up to 30\%, after which sparsity gains are minimal. IG maps are clearer for low-to-moderate pruning (10–30\%), but higher pruning again introduces noise, consistent with reduced sparsity.

\section{Evaluation metrics}\label{appendix:metrics}

\subsection{Sparsity} 
To measure the concentration of attribution in the saliency maps, we use the Gini Index~\cite{chalasani2020conciseexplanationsneuralnetworks}. The Gini Index is a measure of statistical dispersion that quantifies inequality. For a saliency map $S$, the Gini Index is calculated on the absolute values of its attribution scores. A score close to 1 indicates a highly sparse map where attribution is concentrated on a few input features (pixels), while a score close to 0 indicates a diffuse map where attribution is spread out evenly.

\begin{equation}\label{eqn:gini}
    G(\phi(\mathbf{x})) = 1 - 2 \sum_{k=1}^d \frac{\phi(\mathbf{x})_{(k)}}{||\phi(\mathbf{x})||_1} \frac{d-k+0.5}{d}
\end{equation}

Here, \( ||\phi(\textbf{x})||_1 \) is the \( L_1 \)-norm of \( \phi(\textbf{x}) \), and \( \phi(\textbf{x})_{(k)} \) denotes the \( k \)-th smallest element in the sorted vector. The Gini index ranges from 0 to 1. 

Sparsity helps evaluate how concentrated the attribution scores are, with higher sparsity leading to more comprehensible and human-friendly explanations.

\subsubsection{Faithfulness} 

To evaluate how accurately a saliency map reflects the model's decision-making process, we use the Remove and Retrain (ROAD) benchmark~\cite{rong22consistent}. ROAD measures the change in model performance after removing the input features identified as most important by the saliency map. Specifically, we remove the top $k\%$ of pixels (ranked by their attribution scores) and replace them with a neutral value (e.g., average of neighbors). We then measure the model's prediction probability for the correct class on the modified image. A larger drop in probability signifies a more faithful explanation, as it indicates that the removed features were indeed critical to the model's original prediction. 

A sharper accuracy drop as features are removed indicates a better explanation, as the most relevant features have a greater impact on model predictions. 

We quantify the ROAD plot by computing the area over perturbation curve (AOPC) score. 

\section{CRAFT}\label{appendix:craft}
To probe model decisions beyond pixel-level saliency, we adopt CRAFT (Concept Recursive Activation Factorization)~\cite{fel2023craft}, a method for discovering and quantifying concepts directly from model activations.

The first step is to select a set of images predicted as a target class $y$, i.e., $C = \{x_i : f(x_i) = y\}$. This ensures that the analysis reflects the model’s internal representation of a class rather than human labels. From each image, localized crops are extracted via a simple crop-and-resize operator, avoiding artifacts introduced by segmentation and inpainting as in ACE \cite{ghorbani2019towards}. These crops form an auxiliary dataset, which is passed through the network to obtain activations.

To extract concepts, CRAFT applies \textbf{Non-negative Matrix Factorization (NMF)} to the activations, decomposing them into a set of \textit{Concept Activation Vectors (CAVs)} $W$ (the “concept bank”) and corresponding coefficients $U$. This factorization expresses each activation as a non-negative linear combination of concepts, yielding semantically interpretable clusters such as canopy textures, sky patches, or suspension lines in our parachute example.

To evaluate the importance of concepts, CRAFT uses Sobol indices \cite{sobol1993sensitivity}. By perturbing concept coefficients and measuring the variance in model outputs, the method quantifies how much each concept (alone or in interaction with others) influences the prediction. High-importance concepts correspond to activation patterns whose removal or alteration substantially changes the model’s output, while low-importance concepts reflect background or unused features.

We use the official implementation available in \url{https://github.com/deel-ai/Craft}.

\newpage 
\section{Concept extraction}\label{appendix:concept}

\begin{figure}[htbp]
    \centering
    \begin{subfigure}[b]{0.40\textwidth}
        \includegraphics[width=\linewidth]{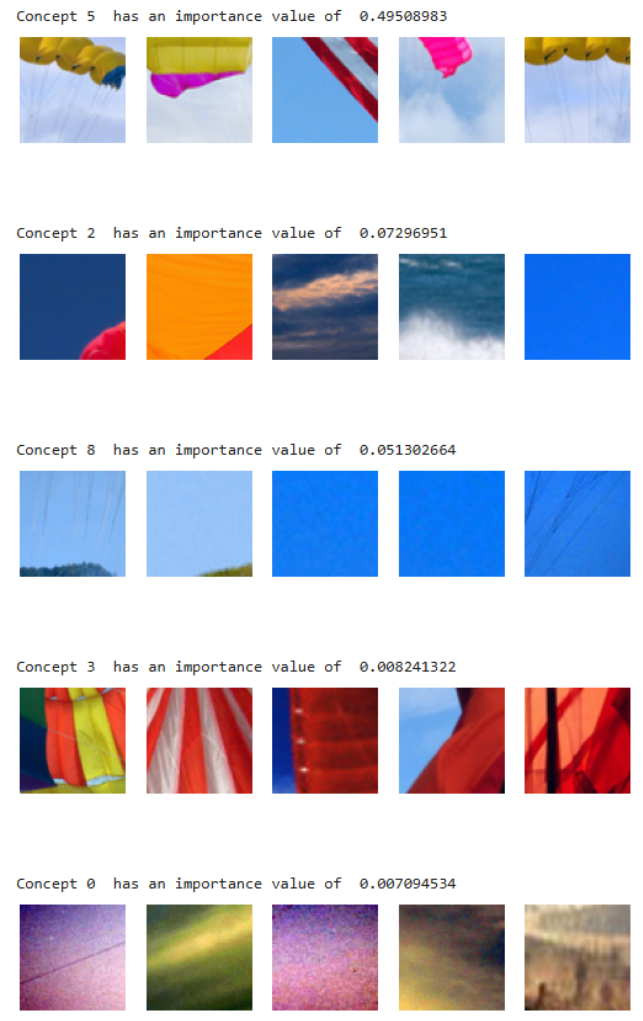}
        \caption{Normal model}
        \label{fig:normalmodel}
    \end{subfigure}
    \hfill % Adds horizontal space between subfigures
    \begin{subfigure}[b]{0.40\textwidth}
        \includegraphics[width=\linewidth]{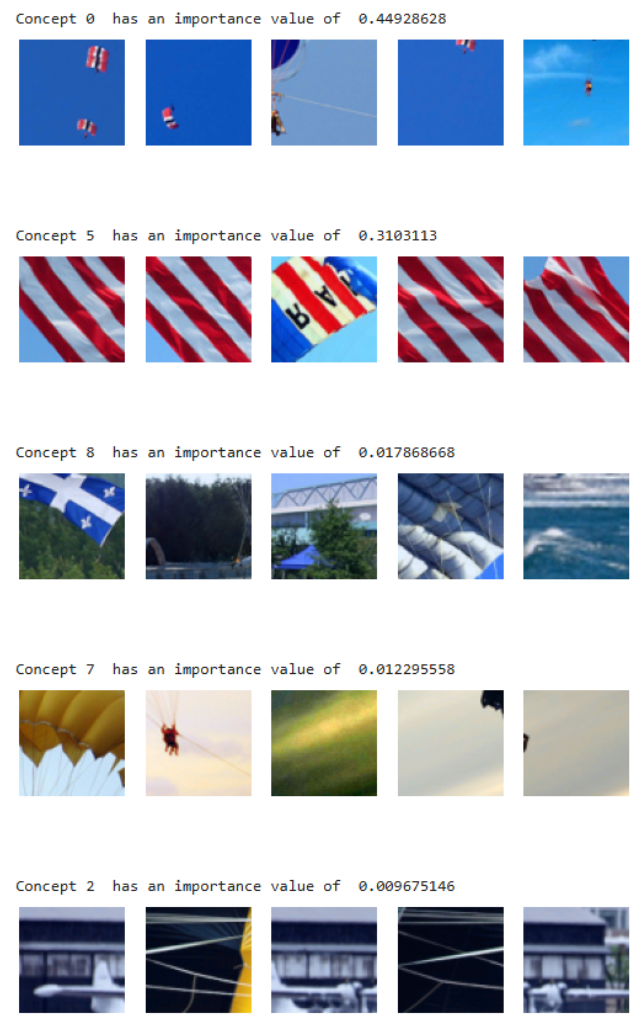}
        \caption{10\% pruned model}
        \label{fig:10model}
    \end{subfigure}
    \caption{Concept extraction}
    \label{fig:concept1}
\end{figure}

\begin{figure}[htbp]
    \centering
    \begin{subfigure}[b]{0.40\textwidth}
        \includegraphics[width=\linewidth]{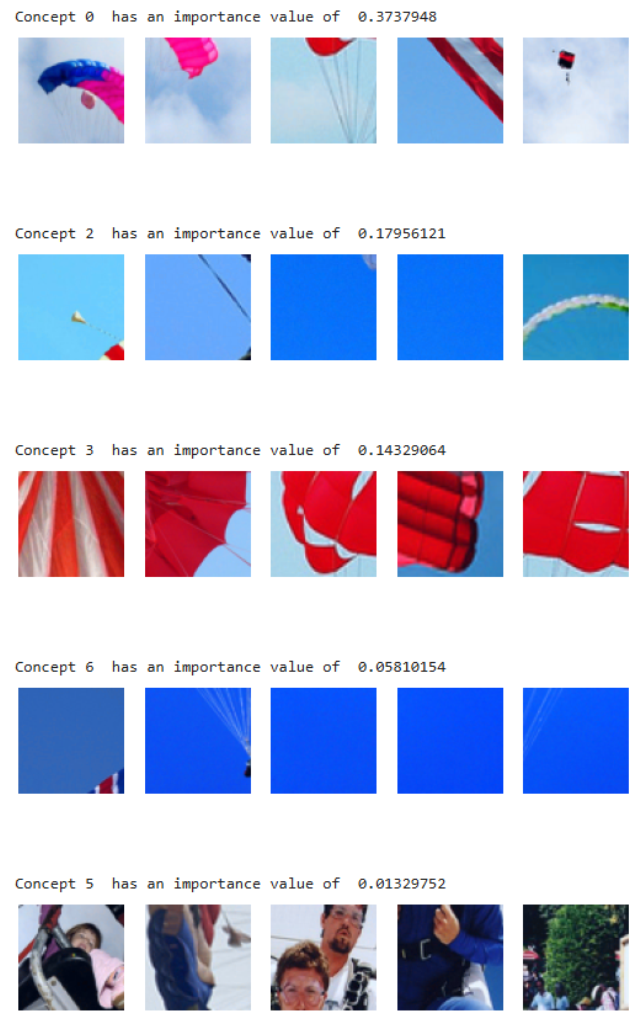}
        \caption{20\% pruned model}
        \label{fig:20model}
    \end{subfigure}
    \hfill % Adds horizontal space between subfigures
    \begin{subfigure}[b]{0.40\textwidth}
        \includegraphics[width=\linewidth]{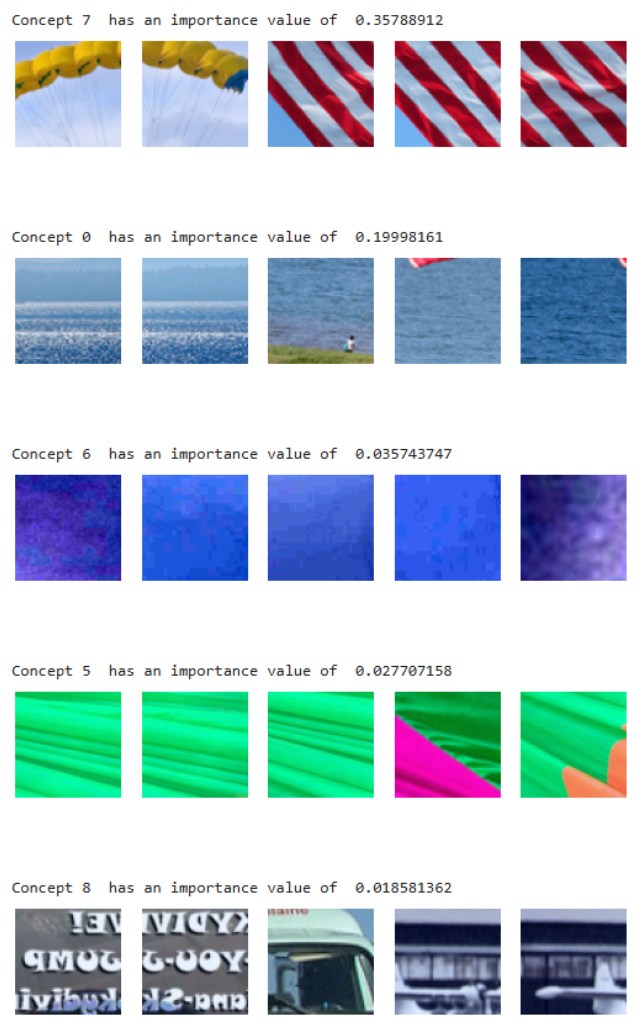}
        \caption{30\% pruned model}
        \label{fig:30model}
    \end{subfigure}
    \caption{Concept extraction}
    \label{fig:concept2}
\end{figure}

\begin{figure}[htbp]
    \centering
    \begin{subfigure}[b]{0.40\textwidth}
        \includegraphics[width=\linewidth]{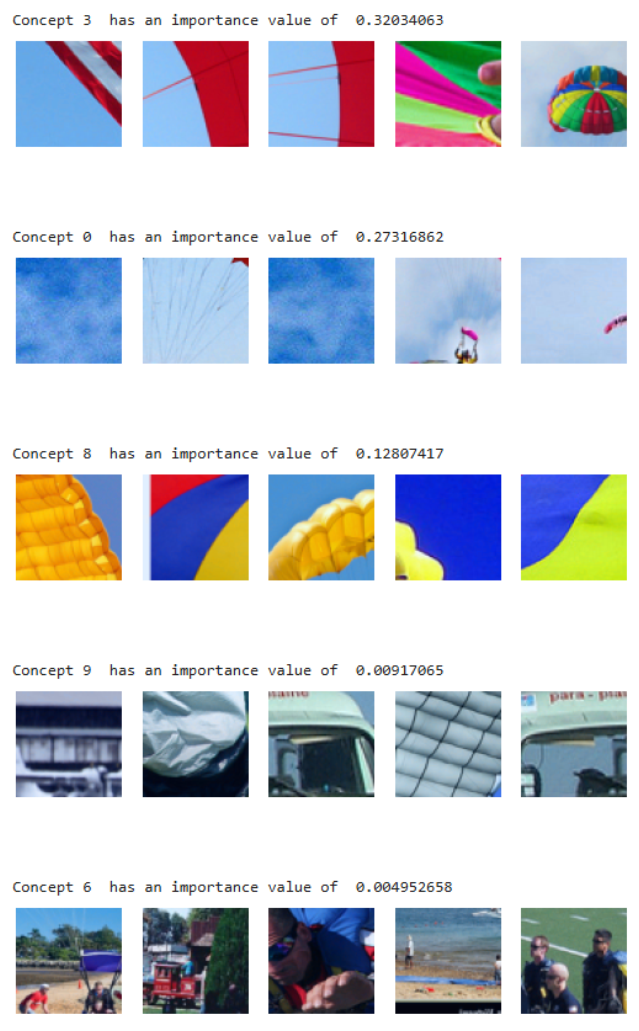}
        \caption{50\% pruned model}
        \label{fig:50model}
    \end{subfigure}
    \hfill % Adds horizontal space between subfigures
    \begin{subfigure}[b]{0.40\textwidth}
        \includegraphics[width=\linewidth]{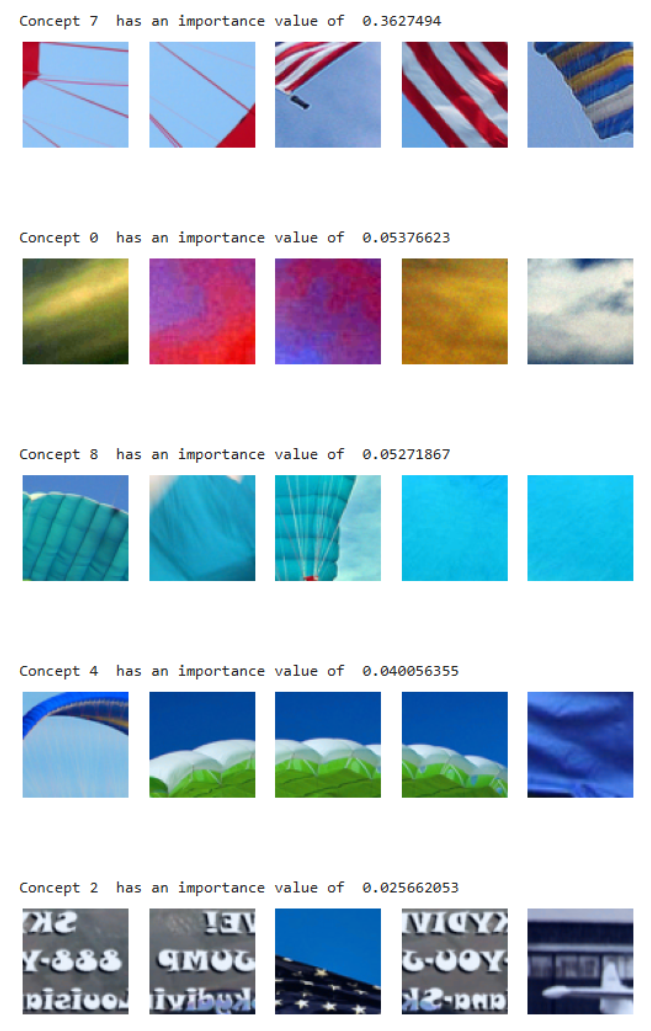}
        \caption{70\% pruned model}
        \label{fig:70model}
    \end{subfigure}
    \caption{Concept extraction}
    \label{fig:concept3}
\end{figure}

Figures \ref{fig:normalmodel}–\ref{fig:70model} illustrate the top CRAFT-extracted concepts for the parachute class across the baseline and progressively pruned models.

For the normal (non-pruned) model, the most influential concept (Concept 5, importance $\approx$ 0.495) captures brightly colored parachute canopies (yellow, magenta, red) against a clear blue sky—highly distinctive for the target class. Secondary concepts include fabric color blocks and environmental elements , sky segments with suspension lines, and multicolored close-ups with lower relevance.

% \begin{figure}[h]
%     \centering
%     \includegraphics[width=0.50\linewidth]{parchuteNormal.png}
%     \caption{Normal model}
%     \label{fig:normalmodel}
% \end{figure}

In the 10\% pruned model, the dominant concept (Concept 0, $\approx$  0.449) shifts toward small parachutes in a clear sky—similar similar the normal model's top concept but more zoomed out and object-centered. The second concept (Concept 5, $\approx$  0.310) emphasizes red-and-white canopy stripes. Lower-ranked concepts include mixed environmental scenes, partial canopy edges with background blur, and structural cord-like patterns.

For the 20\% pruned model, the top concept (Concept 0, $\approx$ 0.374) focuses on close-up canopies in varied colors (blue, pink, red) with sky backgrounds, while Concept 2 ($\approx$ 0.180) captures large blue-sky patches with minimal object detail, indicating greater reliance on background. Concept 3 ($\approx$ 0.143) retains strong red canopy patterns. Concept 6 and Concept 7 include secondary cues like sky textures and even human figures and gear, suggesting increased drift toward non-object-specific features.

In the 30\% pruned model, the leading concept (Concept 7, $\approx$ 0.358) mixes yellow canopy shots with red-and-white patterns, still object-relevant but less semantically pure. Concept 0 ($\approx$ 0.200) shifts to water and shoreline views, marking heavier environmental dependence. Other concepts include abstract blue/purple textures, unrelated green fabric folds, and irrelevant structural or text-based patterns. This stage shows increased background noise infiltration.

By 50\% pruning, the model’s top concepts retain some object cues but lose exclusivity. Concept 3 ($\approx$ 0.320) includes red canopy sections and multicolor parachute views, while Concept 0 ($\approx$ 0.273) largely captures sky backgrounds with partial parachutes. Concept 8 ($\approx$ 0.128) is still object-related, but lower concepts contain unrelated scenes with vehicles, people, and beach environments, reflecting growing concept drift.

The 70\% pruned model relies heavily on a single dominant concept (Concept 7, $\approx$ 0.363) that contains suspension lines, red/white stripes, and yellow-blue canopy shots. Subsequent concepts drop sharply in importance and often mix in non-object patterns: colored textures, blue canopy folds and sky areas, mixed parachute/fabric views, and unrelated text or structural imagery. At this stage, semantic coherence is low outside the top concept.

Overall, these results indicate that low-to-moderate pruning ($\leq$20\%) preserves core parachute concepts while slightly reweighting toward object-centered or background-focused cues. Beyond 30\%, the top concepts increasingly mix heterogeneous patterns, and unrelated or noisy concepts rise in importance. By 70\% pruning, concept purity degrades substantially, suggesting that aggressive sparsification forces the model to compress diverse discriminative features into fewer, less distinct activation patterns, impairing interpretability.

% \begin{figure}[h]
%     \centering
%     \includegraphics[width=0.80\linewidth]{pruned20.png}
%     \caption{20\% pruned model}
%     \label{fig:20model}
% \end{figure}

% \begin{figure}[h]
%     \centering
%     \includegraphics[width=0.80\linewidth]{pruned30.png}
%     \caption{30\% pruned model}
%     \label{fig:30model}
% \end{figure}

% \begin{figure}[h]
%     \centering
%     \includegraphics[width=0.80\linewidth]{pruned50.png}
%     \caption{50\% pruned model}
%     \label{fig:50model}
% \end{figure}

% \begin{figure}[h]
%     \centering
%     \includegraphics[width=0.80\linewidth]{pruned70.png}
%     \caption{70\% pruned model}
%     \label{fig:70model}
% \end{figure}

\end{document}